\documentclass{article}
\usepackage{amssymb, amsmath, latexsym}
\usepackage[dvips]{epsfig}
\newtheorem{theorem}{Th\'eor\`eme}
\newtheorem{lemma}{Lemme}
\newtheorem{proposition}{Proposition}
\newtheorem{corollary}{Corollaire}
\newtheorem{definition}{D\'efinition}

\newtheorem{remark}{Remarque}
\newenvironment{resume}{\noindent \begin{center}  \textbf{R\'esum\'e}  \end{center}  \indent} {\\}
\begin{document}    
\title{\bf R\'eseaux d'Automates de Caianiello Revisit\'e}
\author{Ren\'{e} Ndoundam,  Maurice Tchuente   \\
 {\small  D\'epartement  d'Informatique, } \\
{\small Universit\'e de Yaound\'e I, B.P. 812 Yaound\'e, Cameroun } \\
{\small E.mail: ndoundam@uycdc.uninet.cm, tchuente@uycdc.uninet.cm} \\
       }
\date{}
\maketitle { }
\begin{resume} Nous exhibons une famille d'automates de McCulloch et Pitts 
 de taille $2nk+2$ qui est simulable par un r\'eseau d'automates de Caianiello de taille $2n+2$ et de taille m\'emoire $k$.
Cette simulation nous permet de retrouver l'un des r\'esultats de l'article suivant: [Cycles exponentiels des r\'eseaux de Caianiello et
compteurs en arithm\'etique redondante, Technique et Science Informatiques Vol. 19, pages 985-1008] sur l'existence d'un r\'eseau d'automates de 
Caianiello de taille $2n+2$ et de taille m\'emoire $k$ qui d\'ecrit un cycle de longueur $k \times 2^{nk}$.
\end{resume}
{\bf Mots Cl\'es:} R\'eseau d'automates de Caianiello, m\'emoire, transitoire, cycle.
\section{Introduction}  
Le cerveau humain peut \^etre vu comme un ensemble de neurones interconnect\'es. McCulloch et Pitts ont sugg\'er\'e en 1943 de 
mod\'eliser le cerveau par un r\'eseau d'automates \`a seuil, o\`u chaque automate repr\'esente un neurone formel \cite{MP 43}.

Si nous supposons qu'\`a chaque neurone formel est associ\'e un num\'ero permettant de le r\'ef\'erencer, alors l'\'etat \`a l'instant 
 $t$ du neurone $i$ est donn\'e par:
\begin{equation}      \label{L:num1}
x_i(t+1) = {\bf 1} \Big( \sum_{j=1}^{n}  a_{ij} x_{j}(t) - \theta_i \Big) \ \ \ \ 1 \ \leq i \ \leq n     
\end{equation}
o\`u :
\begin{itemize}
\item[$\bullet$] ${\bf 1}$ est la fonction de Heaviside d\'efinie par:
\[
    {\bf 1}(x) =
      \begin{cases}
      0    &\text{si  $x < 0$  }      \notag      \\
      1    &\text{si  $x \geq 0$ }   \notag
      \end{cases}
\]
\item[$\bullet$] $x_{i}(t)$ est la variable bool\'eenne qui repr\'esente l'\'etat du neurone $i$ au temps
$t$, $x_i(t) \ \in \ \{ 0, 1 \}.$
\item[$\bullet$] $\theta_i$ est le seuil du neurone $i$.
\item[$\bullet$] $a_{ij}$ est un r\'eel appel\'e coefficient d'interaction qui repr\'esente l'influence du neurone $j$ au temps $t$ sur l'\'etat
 du neurone $i$ au temps $t+1$
\item[$\bullet$] $n$ est le nombre de neurones.
\end{itemize}
L'influence du neurone $j$ sur le neurone $i$ est dite excitatrice si $a_{ij}$ est positif, inhibitrice si $a_{ij}$ est n\'egatif, et nulle
si  $a_{ij}$ est \'egal \`a z\'ero. \\

Caianiello et De Luca \cite{CL 66} ont propos\'e l'\'etude de la dynamique de l'\'equation (\ref{L:num1}) dans le cas d'un neurone qui 
n'interagit pas avec aucun autre neurone par une \'equation neuronale de la forme: 

\begin{equation}      \label{L:num2}
x(n) = {\bf 1} \Big( \sum_{j=1}^{k}  a_{j} x(n-j) - \theta \Big)
\end{equation}
o\`u :
\begin{itemize}
\item[$\bullet$] $x(n-j)$ est l'\'etat du neurone \`a l'instant $n-j$.
\item[$\bullet$] $a_j$ est un nombre r\'eel qui repr\'esente l'influence de l'\'etat du neurone \`a l'instant $n-j$ sur l'\'etat du neurone
 \`a l'instant $n$.
\item[$\bullet$] $\theta$ est le seuil d'excitation du neurone. 
\item[$\bullet$] $k$ est la taille de la m\'emoire, c'est-\`a-dire que l'\'etat du neurone \`a l'instant $n$ ne d\'epend que de ses 
 \'etats aux $k$ \'etapes pr\'ec\'edentes. 
\end{itemize}

Le comportement dynamique de l'\'equation neuronale (\ref{L:num2}) ci-dessus est enti\`erement d\'etermin\'e par la taille de la m\'emoire $k$,
le seuil d'excitation $\theta$, les coefficients d'interaction $(a_j)_{1 \leq j \leq k}$ et les $k$ termes initiaux $x_0, x_1, \dots, x_{k-1}$. \\

Caianiello a g\'en\'eralis\'e l'\'etude du r\'eseau neuronal (\ref{L:num1}) de McCulloch et Pitts en consid\'erant que chaque neurone a une taille
m\'emoire $k$

\begin{equation}      \label{L:num3}
y_i(t+1) = {\bf 1} \Big( \sum_{j=1}^{n} \sum_{s=1}^{k} a_{ij}(s) y_{j}(t+1-s) - \theta_i \Big) \ \ \ \ 1 \ \leq i \ \leq n , \ \ t \geq k-1     
\end{equation}
o\`u :
\begin{itemize}
\item[$\bullet$] $y_j(t+1-s)$ est l'\'etat du neurone $j$ \`a l'instant $t+1-s$.
\item[$\bullet$] $a_{ij}(s)$ est un r\'eel qui repr\'esente l'influence de l'\'etat du neurone $j$ \`a l'instant $t+1-s$ sur l'\'etat du neurone
 $i$ \`a l'instant $t+1$.
\item[$\bullet$] $\theta_i$ est le seuil d'excitation du neurone $i$. 
\item[$\bullet$] $\sum_{j=1}^{n} \sum_{s=1}^{k} a_{ij}(s) y_{j}(t+1-s)$ est le potentiel membranaire du neurone $i$ \`a l'instant $t$.
\item[$\bullet$] $n$ est le nombre de neurones du r\'eseau. 
\item[$\bullet$] $k$ est la taille de la m\'emoire. 
\end{itemize}

Dans l'\'etude de la dynamique du r\'eseau neuronal (\ref{L:num1}) de McCulloch et Pitts, Goles et Olivos \cite{GO 81} ont d\'emontr\'e que dans le cas
 o\`u $A=(a_{ij})$ est sym\'etrique, le r\'eseau converge vers un cycle de longueur 2 ou vers un point fixe. Dans l'\'etude du r\'eseau neuronal de 
McCulloch et Pitts, Goles \cite{Gol 82} a d\'emontr\'e que si $A=(a_{ij})$ est sym\'etrique et les \'el\'ements diagonaux sont positifs, l'it\'eration
s\'equentielle converge vers des points fixes. Dans le cas o\`u la matrice $A$ est non sym\'etrique,
Matamala \cite{Mat 95} a exhib\'e une \'evolution cyclique de longueur $2^n$, Ndoundam \cite{Nd 95, NT 00c} a exhib\'e une \'evolution cyclique de
 longueur $2^{n/2}$ o\`u $n$ est le nombre de neurones. Goles et Mart\'{\i}nez \cite{GM 89} ont exhib\'e dans le cas o\`u la matrice d'interaction
 $A=(a_{ij})$ est sym\'etrique une configuration qui d\'ecrit un transitoire de longueur $2^{n/3}$. \\

Dans l'\'etude de la dynamique de l'\'equation neuronale r\'ecurrente (\ref{L:num2}) de Caianiello et De Luca, nous notons $LP(k)$ la plus longue
 p\'eriode des suites que peut g\'en\'erer un neurone dont la taille m\'emoire est $k$. Dans \cite{CMG 88}, il est conjectur\'e que $LP(k)$ est
 inf\'erieure ou \'egale \`a $2k$. Cette conjecture a \'et\'e infirm\'ee dans \cite{Mou 89, CTT 92, TT 93} o\`u on a exhib\'e respectivement des suites
 de p\'eriodes $2k+6$, $O(k^3)$, et $O(e^{\sqrt{k}})$ g\'en\'er\'es par des \'equations neuronales de taille m\'emoire $k$.
Cette borne est port\'ee ais\'ement \`a $O(e^{\sqrt{klog(k)}})$ en prenant la m\^eme construction que dans \cite{TT 93}, mais en se restreignant
 aux p\'eriodes qui sont les nombres premiers.  \\

Dans l'\'etude de la dynamique du r\'eseau neuronal (\ref{L:num3}) de Caianiello, Goles \cite{Gol 85} a d\'emontr\'e que si la classe des matrices
d'interaction est palindromique (c'est-\`a-dire $A(k+1-s) = {}^tA(s)$), alors la p\'eriode $T$ est un diviseur de $k+1$ et d'autre part
 \cite{Gol 80} que si les matrices $A(i), i=1, \dots, k$ ne sont pas sym\'etriques, il existe des cycles de longueur $\frac{(k-1)(k+2)}{2}+1$.  \\

Dans cet article, nous rappelons la simulation d'un compteur BS par un r\'eseau neuronal de McCulloch et Pitts donn\'ee dans \cite{NT 00c}. 
Nous montrons comment une famille de r\'eseaux de McCulloch et Pitts de taille $2nk+2$ peut \^etre simul\'ee par un r\'eseau d'automates de
Caianiello de taille $2n+2$ et de taille m\'emoire $k$. Cette simulation nous permet de retrouver l'un des r\'esultats de l'article \cite{NT 00c}
sur l'existence d'un r\'eseau de neuronal de Caianiello de taille $2n+2$ et de taille m\'emoire $k$ qui d\'ecrit un cycle de longueur $k \times 2^{nk}$. \\

Le reste du papier est organis\'e ainsi: le paragraphe 2 est consacr\'e  \`a la simulation du compteur BS par un r\'eseau neuronal de McCulloch
et Pitts. Le paragraphe 3 pr\'esente les suites $U_n$ et $V_n$. Le paragraphe 4 est dedi\'e la simulation de la suite $V_n$ par un r\'eseau 
neuronal de McCulloch et Pitts. L'\'etude de l'\'evolution pour les r\'eseaux g\'en\'eraux de Caianiello est faite dans le paragraphe 5.
Le paragraphe 6 est consacr\'e \`a la conclusion.

\section{Incr\'ementeur BS}

La notation BS de l'anglais {\it Borrow-Save} par analogie avec la notation {\it Carry-Save} est due \`a Guyot, Herreros et Muller
 \cite{GHM 89}, elle utilise deux bits pour l'\'ecriture d'un chiffre appartenant \`a $\{ -1, 0, 1 \}$. Les nombres sont \'ecrits en base 
 $2$ avec des chiffres \'egaux \`a $\overline{1}$ (pour -1), 0 et 1. On convient de repr\'esenter un chiffre $c$, $c \in \{ -1, 0, 1 \}$   
par un couple de bits $(c^+, c^-)$ tel que $c$ est \'egal \`a $c^+ - c^-$. En notation BS, le code suivant
 $( (a^{+}_{n-1}, a^{-}_{n-1}) \dots (a^{+}_{i}, a^{-}_{i}) \dots (a^{+}_{0}, a^{-}_{0}))$ est \'egal \`a
 $\sum_{i=0}^{n-1} (a^{+}_{i} - a^{-}_{i}) \times 2^i$ en base dix. \\

L'incr\'ementeur BS \cite{DM 91}  utilise le principe de l'additionneur BS introduit par Duprat et Muller \cite{DM 91} et n\'ecessite une seule
 rang\'ee de cellules. Plus pr\'ecis\'ement, soient:  \\
\begin{align}
A \ &= \ ((a^{+}_{n-1}, a^{-}_{n-1}) \dots (a^{+}_{i}, a^{-}_{i}) \dots (a^{+}_{0}, a^{-}_{0}))   \notag   \\
S \ &= \ ((s^{+}_{n}, s^{-}_{n})(s^{+}_{n-1}, s^{-}_{n-1}) \dots (s^{+}_{i}, s^{-}_{i}) \dots (s^{+}_{0}, s^{-}_{0}))   \notag
\end{align} 
o\`u $1$ est cod\'e sous la forme  $((0, 0) \dots (0, 0) (1, 0))$. \\

\begin{figure}[h]
\begin{center}
\scalebox{.75}{\epsfig{file=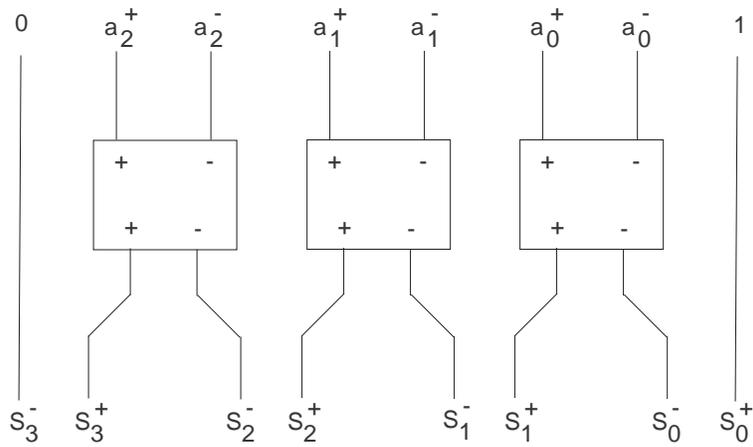}}
\end{center}
\caption{\label{dessin1}{\rm Incr\'ementeur BS} }
\end{figure}

L'incr\'ementeur BS present\'e dans la figure \ref{dessin1} re\c{c}oit en entr\'ee $a^{+}_{i}, \ a^{-}_{i}$ ($0 \leq \ i \ \leq n-1$) et d\'elivre
 en sortie les termes $s^{+}_{i}, \ s^{-}_{i}$ ($0 \leq \ i \ \leq n$).  \\

\bigskip

La cellule \'el\'ementaire d'un incr\'ementeur BS fonctionne comme l'indique la figure \ref{dessin2}: 

\newpage

\begin{figure}[h]
\begin{center}
\scalebox{.75}{\epsfig{file=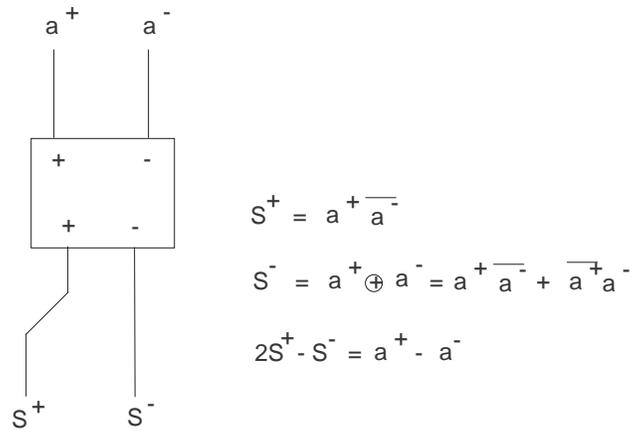}}
\end{center}
\caption{\label{dessin2}{\rm Cellule \'el\'ementaire d'incr\'ementation BS}} 
\end{figure}

En utilisant le fait que $a^{+} \oplus a^{-} \ = \ a^{+}  \overline{a^{-}} + \overline{a^{+}} a^{-}$, l'incr\'ementeur BS se d\'ecompose ais\'ement
en deux \'etages comme present\'e dans la figure \ref{dessin3}: 

\newpage

\begin{figure}[h]
\begin{center}
\scalebox{.75}{\epsfig{file=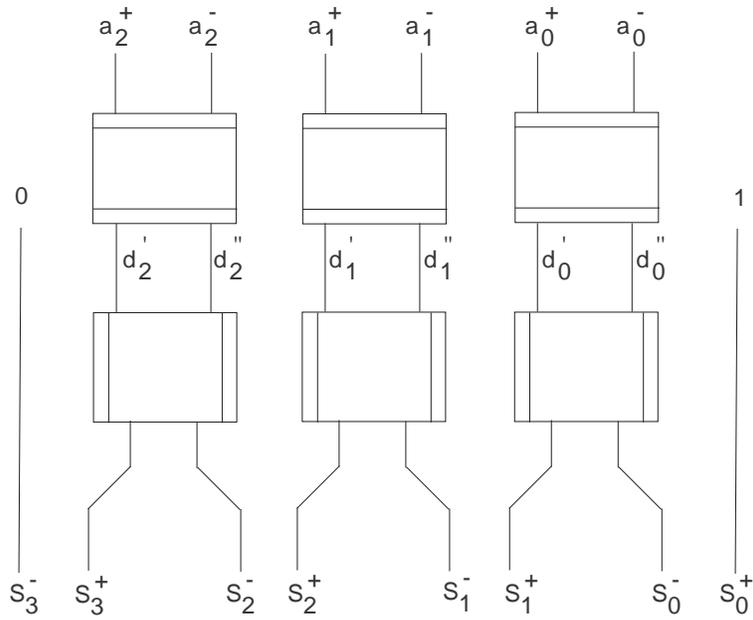}}
\end{center}
\caption{\label{dessin3}{\rm Incr\'ementeur BS \`a deux \'etages}}
\end{figure}

{\bf Premi\`ere \'etape:}  \\
Le premier \'etage re\c{c}oit en entr\'ee $((a^{+}_{n-1}, a^{-}_{n-1}) \dots (a^{+}_{i}, a^{-}_{i}) \dots (a^{+}_{0}, a^{-}_{0}))$, calcule les termes
$a^{+}_{i} \overline{a^{-}_{i}}$, $\overline{a^{+}_{i}} a^{-}_{i}$ et d\'elivre en sortie $(d^{'}_{n-1}, d^{''}_{n-1}, \dots, d^{'}_{0}, d^{''}_{0})$  \\
selon les formules:
\[
d^{'}_{i} \ = \ a^{+}_{i} \overline{a^{-}_{i}} \ \ {\rm \ et \  } \ d^{''}_{i} \ = \ \overline{a^{+}_{i}} a^{-}_{i} \ \ \ \ 0 \ \leq \ i \ \leq \ n-1
\]
{\bf Deuxi\`eme \'etape:} \\
 
Le second \'etage re\c{c}oit en entr\'ee $(d^{'}_{n-1}, d^{''}_{n-1}, \dots, d^{'}_{0}, d^{''}_{0})$ et d\'elivre en sortie 
 $((s^{+}_{n}, s^{-}_{n})  \dots (s^{+}_{i}, s^{-}_{i}) \dots (s^{+}_{0}, s^{-}_{0}))$ selon les formules: \\

\begin{align}   
s^-_i \ &= \ d^{'}_{i} + d^{''}_{i}   \ &\ 0 \ \leq \ i \ \leq \ n-1   \notag  \\
s^+_i \ &= \ d^{'}_{i}                \ &\ 0 \ \leq \ i \ \leq \ n-1   \notag
\end{align}

avec les conditions aux bords suivantes:
\[
s^+_0 \ = \ 1  \ \ \ {\rm \ et \ } s^-_n \ = \ 0
\]

\section{Suites $U_n$ et $V_n$}

Ce paragraphe est consacr\'e \`a l'\'etude de deux suites construites \`a partir de l'incr\'ementeur BS. Le sous-paragraphe 3.1 est dedi\'e \`a
 la pr\'esentation de la suite $U_n$, le sous-paragraphe 3.2 pr\'esente la construction de la suite $V_n$. Le sous-paragraphe 3.3 est consacr\'e
 \`a la relation de passage entre deux termes cons\'ecutifs de la suite $V_n$.

\subsection{Pr\'esentation de la suite $U_n$}

La notation BS est redondante, c'est-\`a-dire un nombre peut avoir plusieurs 
repr\'esentations (ou codes) possibles, ceci nous conduit \`a introduire la notion d'identit\'e de code:

\begin{definition}
{\rm Soient deux codes} $A_{BS}$ {\rm et} $B_{BS}$ {\rm de deux nombres} $A$ {\rm et} $B$ {\rm \'ecrits en notation BS,}
\begin{align}     \notag
 A_{BS}  &=  ( (a^{+}_{n-1}, a^{-}_{n-1}) \dots (a^{+}_{i}, a^{-}_{i}) \dots (a^{+}_{0}, a^{-}_{0}))   \notag   \\
 B_{BS}  &=  ( (b^{+}_{n-1}, b^{-}_{n-1}) \dots (b^{+}_{i}, b^{-}_{i}) \dots (b^{+}_{0}, b^{-}_{0}))   \notag   
\end{align}
$A_{BS}$ {\rm et } $B_{BS}$ {\rm sont identiques si et seulement si:}
\[
a^{+}_{j} \ = \ b^{+}_{j} \  {\rm et} \ a^{-}_{j} \ = b^{-}_{j}  \  \  \  , \ 0 \ \leq j \ \leq n-1 
\]
\end{definition}

{\bf Commentaire 1.} A cause de la redondance du syst\`eme BS, deux nombres \'ecrits en notation BS peuvent \^etre \'egaux sans \^etre
identique.  \\

Partant de l'incr\'ementeur BS \`a deux \'etages, on introduit une suite $U_n$ telle que $U_n(k)$ soit un repr\'esentant de $k$ en 
notation BS.

\begin{definition} {\rm Suite} $U_n$  \label{L:def1}   \\
$\forall \ j, n \in \mathbb{N}$, $n \geq 3$ \\
$U_n(0)$ {\rm est identique \`a} $( (a^{+}_{n-1}, a^{-}_{n-1}) \dots (a^{+}_{0}, a^{-}_{0}))$ 
\[
   {\rm \ o\grave{u} \ }
   \begin{cases}
        a^{+}_{i} =  a^{-}_{i} = 1   &\text{si $i \in \{0, 1 \}$}     \notag  \\
        a^{+}_{i} =  a^{-}_{i} = 0   &\text{sinon}     \notag  
    \end{cases}
\]

$U_n(j+1)$ {\rm est identique \`a} $((d^{+}_{n-1}, d^{-}_{n-1}) \dots (d^{+}_{0}, d^{-}_{0}))$ \\  {\rm avec}    
 $((d^{+}_{n}, d^{-}_{n})(d^{+}_{n-1}, d^{-}_{n-1}) \dots (d^{+}_{0}, d^{-}_{0})) \ = \ U_n(j)+1$ {\rm o\`u}  $1$ {\rm est cod\'e sous la forme}
 $((0, 0) \dots (0, 0) (1, 0))$ 
\end{definition}
\begin{remark}
Pour obtenir $U_n(j+1)$, on incr\'emente $U_n(j)$ gr\^ace \`a l'incr\'ementeur BS, mais on ne retient que les 
 $2n$ bits de poids faibles, car on travaille modulo $2^n$.  
\end{remark}

En se servant de la notion d'identit\'e de code en notation BS, on \'etablit les r\'esultats suivants:

\begin{lemma}  \label{LL:lem1}
$\forall \ i, n \in \mathbb{N}$, $n \geq 3$  \\
{\rm Si:} 
\begin{align}
&U_n(i)     &{\rm \ est \ identique \ \grave{a}} \ \ &((a^{+}_{n-1}, a^{-}_{n-1}) \dots (a^{+}_{1}, a^{-}_{1})(a^{+}_{0}, a^{-}_{0}))  \notag  \\
&U_n(i+1)   &{\rm \ est \ identique \ \grave{a}} \ \ &((b^{+}_{n-1}, b^{-}_{n-1}) \dots (b^{+}_{1}, b^{-}_{1})(b^{+}_{0}, b^{-}_{0}))  \notag   \\
&U_{n+1}(i) &{\rm \ est \ identique \ \grave{a}} \ \ &((c^{+}_{n}, c^{-}_{n})(c^{+}_{n-1}, c^{-}_{n-1}) \dots (c^{+}_{1}, c^{-}_{1})(c^{+}_{0}, c^{-}_{0}))    \notag 
\end{align}
{\rm Alors,}       \\
$U_{n+1}(i+1)$ {\rm est identique \`a} 
$((a^{+}_{n-1} \overline{a^{-}_{n}} , c^{+}_{n} \oplus c^{-}_{n}) (b^{+}_{n-1}, b^{-}_{n-1}) \dots (b^{+}_{1}, b^{-}_{1})
          (1 , b^{-}_{0}))$  
\end{lemma}
{\bf D\'emonstration.} Posons $a^{+}_{0} = b^{+}_{0} = c^{+}_{0} = 1$, par d\'efinition des termes g\'en\'eraux des suites $U_n$, $U_{n+1}$
et du principe de l'incr\'ementeur BS, on a l'\'egalit\'e suivante:    
\begin{equation}      \label{L:num4}
\forall \ j \in \mathbb{N}, \ 0 \leq j \leq n-1 \ \ \ a^{+}_{j} = c^{+}_{j} \ \ {\rm et} \ \ a^{-}_{j} = c^{-}_{j}
\end{equation}
Du principe de l'incr\'ementeur BS et de l'\'equation (\ref{L:num4}), on d\'eduit:
\begin{equation}      \label{L:num5}
\forall \ j \in \mathbb{N}, \ 0 \leq j \leq n-1 \ \ \ a^{+}_{j} \oplus a^{-}_{j} \ = \ c^{+}_{j} \oplus c^{-}_{j} \ = \ b^{-}_{j} 
\end{equation}
\begin{equation}      \label{L:num6}
\forall \ j \in \mathbb{N}, \ 0 \leq j \leq n-2 \ \ \ a^{+}_{j} \overline{a^{-}_{j}}  \ = \ c^{+}_{j} \overline{c^{-}_{j}} \ = \ b^{+}_{j+1} 
\end{equation}
De la d\'efinition \ref{L:def1}, on d\'eduit que: $U_{n+1}(i+1)$ est identique \`a    \\
$( (c^{+}_{n-1} \overline{c^{-}_{n-1}} , c^{+}_{n} \oplus c^{-}_{n}) (c^{+}_{n-2} \overline{c^{-}_{n-2}} , c^{+}_{n-1} \oplus c^{-}_{n-1})
    \dots  (c^{+}_{0} \overline{c^{-}_{0}} , c^{+}_{1} \oplus c^{-}_{1}) (1 , c^{+}_{0} \oplus c^{-}_{0}))$  \\
Des \'equations (\ref{L:num4}), (\ref{L:num5}) et (\ref{L:num6}), on d\'eduit que $U_{n+1}(i+1)$ est identique \`a   \\ 
$((a^{+}_{n-1} \overline{a^{-}_{n-1}} , c^{+}_{n} \oplus c^{-}_{n}) (b^{+}_{n-1}, b^{-}_{n-1}) \dots (b^{+}_{1}, b^{-}_{1})
          (1 , b^{-}_{0}))$.  \\
$\blacksquare$  

\begin{lemma}   \label{LL:lem2}   
$\forall \ k, n \in \mathbb{N}$, $n \geq 3$   \\
{\rm Si:}  
\begin{equation}    \notag
U_n(k) \ = \ ( (a^{+}_{n-1}, a^{-}_{n-1}) \dots (a^{+}_{0}, a^{-}_{0}))   
\end{equation}  
{\rm Alors,}  
\begin{equation}    \notag
\sum_{i=0}^{n-1} ( a^{+}_{i} - a^{-}_{i} )  \times 2^i  \ \equiv k \Big[ 2^n \Big]
\end{equation}  
\end{lemma}
{\bf D\'emonstration.} Suivant les valeurs de $k$, on distingue deux cas:  \\
{\bf Premier Cas:}  \ \ $0 \ \leq k \ \leq \ n$   \\
Par d\'efinition de $U_n(k)$, il est imm\'ediat que: $\sum_{i=0}^{n-1} ( a^{+}_{i} - a^{-}_{i} )  \times 2^i \ = \ k$ \\
{\bf Deuxi\`eme  Cas:}  \ \ $n \ < \ k$   \\
Soit la suite $A$ \'ecrite en notation BS, dont le terme g\'en\'eral est d\'efini comme suit: 
\begin{center}
$A(n)$ est identique \`a $U_n(n)$. 
\end{center}
Pour tout entier naturel $k$ sup\'erieur ou \'egal \`a $n$, $A(k+1)$ est d\'efini \`a partir de
 $A(k) = ((d^{+}_{k-1}, d^{-}_{k-1}) \dots (d^{+}_{0}, d^{-}_{0}))$ comme suit:  
\begin{align}
A(k+1) \ &= \ ( (e^{+}_{k}, e^{-}_{k}) \dots (e^{+}_{0}, e^{-}_{0}))       \notag   \\
         &= \ A(k)+ 1 \ \ o\grave{u} \ {\rm 1 \ est \ cod\acute{e} \ sous \ la \ forme} \ ((0, 0) \dots (0, 0)(1, 0))  \notag \\
         &=  ((d^{+}_{k-1}, d^{-}_{k-1}) \dots (d^{+}_{0}, d^{-}_{0})) + ((0, 0) \dots (0, 0)(1, 0))   \notag
\end{align}
$A(k+1)$ est le r\'esultat de l'incr\'ementation de $A(k)$ gr\^ace \`a l'incr\'ementeur BS, il est imm\'ediat que:
\begin{align}
k \ &= \ \sum_{i=0}^{k-1}  (d^+_i - d^-_i) \times 2^i     \notag   \\
  \ &= \ \sum_{i=0}^{n-1}  (d^+_i - d^-_i) \times 2^i + \sum_{i=0}^{k-1-n}  (d^+_{n+i} - d^-_{n+i}) \times 2^{n+i} \notag \\     
 \ &= \ \sum_{i=0}^{n-1}  (d^+_i - d^-_i) \times 2^i + 2^n \Bigg( \sum_{i=0}^{k-1-n}  (d^+_{n+i} - d^-_{n+i}) \times 2^{i} \Bigg) \notag 
\end{align}
Or $\forall \ i \in \mathbb{N}$, $0 \ \leq \ i \ \leq n-1$,  $a^+_i \ = \ d^+_i$  et  $a^-_i \ = \ d^-_i$. Il s'ensuit que:
\begin{equation} 
\sum_{i=0}^{n-1} ( a^{+}_{n-1} - a^{-}_{n-1} ) \times 2^i \ \equiv k \Big[ 2^n \Big]
\end{equation}
$\blacksquare$  

\begin{lemma}   \label{LL:lem3}   
$\forall \ n \in \mathbb{N}$, $n \geq 3$, $U_n(2^n)$ est identique \`a $U_n(0)$  
\end{lemma}
{\bf D\'emonstration.} \\
On fera la d\'emonstration par r\'ecurrence sur $n$ \`a partir du rang 3, \\
{\bf Cas de base.}  $n=3$   \\
En utilisant la d\'efinition \ref{L:def1}, on v\'erifie ais\'ement que:   
\begin{center}
$U_3(0)$ {\rm est identique \`a} $( (0,0)(1,1)(1,1) )$       \\
$U_3(1)$ {\rm est identique \`a} $( (0,0)(0,0)(1,0) )$       \\
$U_3(2)$ {\rm est identique \`a} $( (0,0)(1,0)(1,1) )$       \\ 
$U_3(3)$ {\rm est identique \`a} $( (1,0)(0,1)(1,0) )$       \\
$U_3(4)$ {\rm est identique \`a} $( (0,1)(1,1)(1,1) )$       \\
$U_3(5)$ {\rm est identique \`a} $( (0,1)(0,0)(1,0) )$       \\
$U_3(6)$ {\rm est identique \`a} $( (0,1)(1,0)(1,1) )$       \\
$U_3(7)$ {\rm est identique \`a} $( (1,1)(0,1)(1,0) )$       \\
$U_3(8)$ {\rm est identique \`a} $( (0,0)(1,1)(1,1) )$       
\end{center}

{\bf Hypoth\`ese de r\'ecurrence.} On suppose la propri\'et\'e vraie au rang $k$,  \\ c'est-\`a-dire: 
\begin{equation}     \label{L:num7}
 U_k(2^k) {\rm \ est \ identique \ \grave{a}} \ \ U_k(0) 
\end{equation}

{\bf Rang k+1.}  Supposons: 
\begin{align}
&U_k(0) {\rm \ est \ identique \ \grave{a}} \ ((a^{+}_{k-1}, a^{-}_{k-1}) \dots (a^{+}_{0}, a^{-}_{0})),  \notag  \\
&U_k(2^k-1) {\rm \ est \ identique \ \grave{a}} \ ((x^{+}_{k-1}, x^{-}_{k-1}) \dots (x^{+}_{0}, x^{-}_{0})),  \notag  \\
&U_k(2^k) {\rm \ est \ identique \ \grave{a}} \ ((y^{+}_{k-1}, y^{-}_{k-1}) \dots (y^{+}_{0}, y^{-}_{0})),  \notag  \\
&U_k(2^{k+1}-1) {\rm \ est \ identique \ \grave{a}} \ ((t^{+}_{k-1}, t^{-}_{k-1}) \dots (t^{+}_{0}, t^{-}_{0})),  \notag  \\
&U_k(2^{k+1}) {\rm \ est \ identique \ \grave{a}} \ ((z^{+}_{k-1}, z^{-}_{k-1}) \dots (z^{+}_{0}, z^{-}_{0})),  \notag  \\
&U_{k+1}(2^{k+1}-1) {\rm \ est \ identique \ \grave{a}} \ ((q^{+}_{k}, q^{-}_{k}) \dots (q^{+}_{0}, q^{-}_{0})),      \notag  \\
&U_{k+1}(2^{k+1}) {\rm \ est \ identique \ \grave{a}} \ ((r^{+}_{k}, r^{-}_{k}) \dots (r^{+}_{0}, r^{-}_{0})).      \notag  
\end{align}  

Du principe de l'incr\'ementeur BS et de l'\'equation (\ref{L:num7}), on d\'eduit
\[
y^{-}_{k-1} \ = \ x^+_{k-1} \oplus  x^-_{k-1} \ = \ 0
\]

Par cons\'equent, on a: 
\begin{equation}   \label{L:num8}
 x^+_{k-1} \ = \ x^-_{k-1}
\end{equation}
L'\'equation (\ref{L:num7}) implique que $U_k(2^{k+1}-1)$ est identique \`a $U_k(2^{k}-1)$, ce qui entra\^ine:
\begin{equation}
\forall \ i \in \mathbb{N} , \ 0 \leq i \leq k-1  \ \ \ t^+_i \ = \ x^+_i \ \ {\rm et} \ \  t^-_i \ = \ x^-_i    \label{L:num9}  
\end{equation}
On a:
\begin{equation}
\forall \ i \in \mathbb{N}, \ 0 \leq i \leq k-1  \ \ \ a^+_i \ = \ a^-_i    \label{L:num10}
\end{equation}
Par application du Lemme \ref{LL:lem1}, on a:
\[
U_{k+1}(2^{k+1}) \ = \ ( ( t^{+}_{k-1}  \overline{t^{-}_{k-1}}, q^{+}_{k} \oplus q^{-}_{k}) ( z^+_{k-1}, z^-_{k-1}) \dots (z^{+}_{0}, z^{-}_{0}))
\] 
Par application du Lemme \ref{LL:lem2}, on a \'egalement:  
\[
( (t^+_{k-1} \overline{t^-_{k-1}}) - ( q^+_{k-1} \oplus q^-_{k-1} )  ) \times 2^k + \ \sum_{i=0}^{k-1} ( z^+_i - z^-_i ) \times 2^i
                       \equiv 0  \Big[ 2^{k+1} \Big]
\]

\begin{remark}
 il y a abus de l'utilisation des expressions bool\'eennes dans les expressions arithm\'etiques. 
\end{remark}
Les \'equations (\ref{L:num8}) et (\ref{L:num9}) entra\^inent que:
\[
t^+_{k-1} \ = \ x^+_{k-1} \ = \ x^-_{k-1} \ = \ t^-_{k-1}   
\]
par cons\'equent  $t^+_{k-1} \overline{t^-_{k-1}} \ = \ 0$.  \\
$U_k(2^{k+1})$ est identique \`a $U_k(2^k)$, et $U_k(2^k)$ est identique \`a $U_k(0)$, par transitivit\'e $U_k(2^{k+1})$ est identique \`a
$U_k(0)$. Ceci entra\^ine que:
\[
\forall \ i \in \mathbb{N}, \ 0 \ \leq \ i \ \leq k-1 \ \ \ z^+_i \ = \ z^-_i
\]
De ce qui pr\'ec\`ede, on d\'eduit:
\[
q^+_k \oplus q^-_k \ = \ 0
\]
Ce qui entra\^ine que:
\[
U_{k+1}(2^{k+1}) \ {\rm \ est \ identique \ \grave{a}} \ U_{k+1}(0).   
\]
$\blacksquare$  

\begin{lemma}   \label{LL:lem4}
$\forall \ n, i, j \in \mathbb{N}, \ n \geq 3 \ , \ \ 0 \ \leq i \ < j \ < \ 2^n$, $U_n(i)$ \ {\rm n'est \ pas \ identique \`a } \ $U_n(j)$.
\end{lemma}
{\bf D\'emonstration.} \\
Supposons: 
\begin{center}
$U_n(i)$ est identique \`a $((a^{+}_{n-1}, a^{-}_{n-1}) (a^{+}_{n-2}, a^{-}_{n-2}) \dots (a^{+}_{0}, a^{-}_{0}))$,  \\
$U_n(j)$ est identique \`a $((b^{+}_{n-1}, b^{-}_{n-1}) (b^{+}_{n-2}, b^{-}_{n-2}) \dots (b^{+}_{0}, b^{-}_{0}))$,  
\end{center}
Par application du Lemme \ref{LL:lem2}, on d\'eduit:

\begin{equation}   \label{L:num11}
   \begin{cases}
        \sum_{k=0}^{n-1}  (a^{+}_{k} -  a^{-}_{k} ) \times 2^k \equiv i \Big[ 2^n \Big]   \\
        \sum_{k=0}^{n-1}  (b^{+}_{k} -  b^{-}_{k} ) \times 2^k \equiv j \Big[ 2^n \Big]  
    \end{cases}
\end{equation}
La d\'emonstration se fera par l'absurde, supposons $U_n(i)$ identique \`a $U_n(j)$. Ceci entra\^ine:
\begin{equation}   \label{L:num12}
a^+_k \ = \ b^+_k \ \ {\rm et} \ \ a^-_k \ = \ b^-_k  \ \  \ \ \ \ 0 \ \leq \ k \ \leq \ n-1
\end{equation}  
Des \'equations (\ref{L:num11}) et (\ref{L:num12}), on d\'eduit $i \equiv j [ 2^n ]$, or par hypoth\`ese $0 \ \leq  i \ < \ j \ < \ 2^n$,
d'o\`u contradiction.  \\
$\blacksquare$ 

\subsection{Pr\'esentation de la suite $V_n$}

En entr\'ee et en sortie de l'incr\'ementeur BS \`a deux \'etages de la figure \ref{dessin3}, $(a^+_i, a^-_i)$ et $(s^+_i, s^-_i)$ repr\'esentent
 respectivement les termes de rang $i$ des nombres \\ $((a^{+}_{n-1}, a^{-}_{n-1}) (a^{+}_{n-2}, a^{-}_{n-2}) \dots (a^{+}_{0}, a^{-}_{0}))$
  et $((s^{+}_{n}, s^{-}_{n}) (s^{+}_{n-1}, s^{-}_{n-1}) \dots (s^{+}_{0}, s^{-}_{0}))$ \'ecrits en notation BS. \\
S\'emantiquement, $d^{'}_i$ et $d^{''}_{i+1}$ ne repr\'esentent pas les bits de rang $i$ d'un nombre \'ecrit en notation BS. \\
Supposons qu'on ait en entr\'ee du premier \'etage de l'incr\'ementeur BS \`a deux \'etages de la figure \ref{dessin3}, le terme $U_n(i)$ 
identique \`a $((a^{+}_{n-1}, a^{-}_{n-1}) (a^{+}_{n-2}, a^{-}_{n-2}) \dots (a^{+}_{0}, a^{-}_{0}))$. Les $2n$ bits de poids faibles
 $((s^{+}_{n-1}, s^{-}_{n-1}) \dots (s^{+}_{0}, s^{-}_{0}))$ du nombre  \\ $((s^{+}_{n}, s^{-}_{n}) (s^{+}_{n-1}, s^{-}_{n-1}) \dots
 (s^{+}_{0}, s^{-}_{0}))$ en sortie du deuxi\`eme \'etage de l'incr\'ementeur BS \`a deux \'etages repr\'esentent le terme de rang
 $i+1$ de la suite $U_n$, c'est-\`a-dire $U_n(i+1)$.  \\
Les termes $d^{'}_i$ et $d^{''}_i$ seront \'egaux respectivement \`a $a^+_i \overline{a^-_i}$ et $\overline{a^+_i} a^-_i$. Nous d\'efinissons
le terme g\'en\'eral de la suite $V_n$ \`a valeurs dans $\{0, 1 \}^{2n+2}$ \`a partir des termes $d^{'}_i$ et $d^{''}_i$.
   
\begin{definition}: {\rm Suite} $V_n$  \label{L:def3}       \\
$\forall \ n, i \ \in \ \mathbb{N}$, $n \geq 3$  \\
{\rm Si:}   
\begin{center}
$U_n(i)$ {\rm est identique \`a} $((a^{+}_{n-1}, a^{-}_{n-1}) (a^{+}_{n-2}, a^{-}_{n-2}) \dots (a^{+}_{0}, a^{-}_{0}))$  
\end{center}
{\rm Alors,}   \\  
$V_n(i+1)$ {\rm est \'egal \`a} $(d^{'}_{n-1}, 0, d^{'}_{n-2},  d^{''}_{n-1}, \dots, d^{'}_{j},  d^{''}_{j+1}, \dots , d^{'}_{0}, d^{''}_{1},
               1, d^{''}_{0} )$  \\
{\rm o\`u } $d^{'}_{j} \ = \ a^+_j \overline{a^-_j}$ {\rm et} $d^{''}_{j} \ = \ \overline{a^+_j} a^-_j$ \ \ \ , \ $0 \ \leq \ j \ \leq \ n-1$
\end{definition}

\subsection{Relation de passage entre deux termes cons\'ecutifs de la suite $V_n$}

Notre objectif est de d\'eriver la relation de passage entre deux termes cons\'ecutifs de la suite $V_n$. Pour cela, consid\'erons
 $i \in \mathbb{N}$, \\
supposons:  
\begin{align}
&U_n(i) {\rm \ est \ identique \ \grave{a}} \ ((a^{+}_{n-1}, a^{-}_{n-1}) (a^{+}_{n-2}, a^{-}_{n-2}) \dots (a^{+}_{0}, a^{-}_{0})) \notag  \\
&U_n(i+1) {\rm \ est \ identique \ \grave{a}} \ ((b^{+}_{n-1}, b^{-}_{n-1}) (b^{+}_{n-2}, b^{-}_{n-2}) \dots (b^{+}_{0}, b^{-}_{0})) \notag  
\end{align}
De m\^eme, supposons que:
\begin{align}
&V_n(i+1)  {\rm \ est \ \acute{e}gal \ \grave{a}} \ (c^{'}_{n}, c^{''}_{n}, \dots, c^{'}_{0}, c^{''}_{0})   \notag  \\
&V_n(i+2)  {\rm \ est \ \acute{e}gal \ \grave{a}} \ (d^{'}_{n}, d^{''}_{n}, \dots, d^{'}_{0}, d^{''}_{0})   \notag
\end{align}

Dans le diagramme suivant, les fl\`eches en traits pleins mat\'erialisent les transformations qui sont connues. Les fl\`eches 
 en traits interrompus d\'esignent les transformations que nous allons d\'eriver. 

\newpage

\begin{figure}[h]
\begin{center}
\scalebox{.75}{\epsfig{file=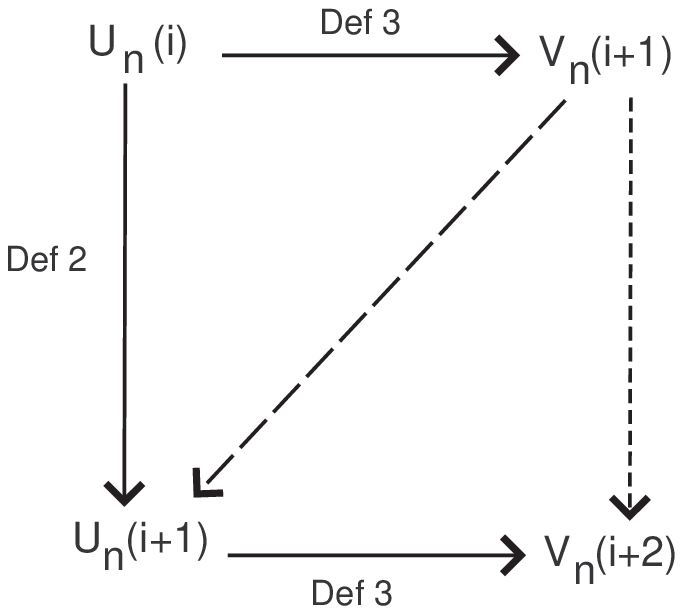}}
\end{center}
\caption{\label{dessin4}{ } }
\end{figure}

De la d\'efinition \ref{L:def3}, on a les conditions aux bords suivantes:
\[
c^{'}_{0} \ = \ 1 \ \ , \ \ c^{''}_{n} \ = \ 0
\]     
et les relations suivantes:
\begin{equation}  \label{L:num14}
c^{'}_{i+1} \ = \ a^+_i \overline{a^-_i} \ \ , \ \ c^{''}_{i} \ = \ \overline{a^+_i} a^-_i \ \ {\rm pour \ tout \ i \ v\acute{e}rifiant} \ 0 \ \leq i \
  \leq n-1  
\end{equation}
De m\^eme, de la d\'efinition \ref{L:def3} on a les conditions aux bords suivantes
\[
d^{'}_{0} \ = \ 1 \ \ , \ \ d^{''}_{n} \ = \ 0
\]    
et les relations suivantes:
\begin{equation}  \label{L:num16}
d^{'}_{i+1} \ = \ b^+_i \overline{b^-_i} \ \ , \ \ d^{''}_{i} \ = \ \overline{b^+_i} b^-_i \ \ {\rm pour \ tout \ i \ v\acute{e}rifiant} \ 0 \ \leq i \
  \leq n-1  
\end{equation}
De la d\'efinition \ref{L:def1}, on a les relations suivantes:
\begin{equation}  \label{L:num18}
   \begin{cases}
        b^{+}_{0} =  1                             &\text{}    \\
        b^{-}_{i} =  a^{+}_{i} \oplus a^{-}_{i}    &\text{, $0 \ \leq i \ \leq n-1$}      \\
        b^{+}_{i+1} =  a^+_i \overline{a^-_i} \    &\text{, $0 \ \leq i \ \leq n-2$}     \\
    \end{cases}
\end{equation}
Notre probl\`eme, c'est d'exprimer les $d^{'}_{i}$, $d^{''}_{i}$ en fonction des $c^{'}_{i}$ et $c^{''}_{i}$. On exprime tout
d'abord les $b^{+}_{i}$, $b^{-}_{i}$ en fonction des $c^{'}_{i}$ et $c^{''}_{i}$.  \\
${\bf 1^{\grave{e}re} Etape:}$  \\
Des \'equations (\ref{L:num14}) et (\ref{L:num18}), on d\'eduit:
\begin{equation}  \label{L:num20}
   \begin{cases}
        b^{+}_{0} =  1                             &\text{}    \\
        b^{-}_{i} =  a^{+}_{i} \oplus a^{-}_{i} \ = \  c^{'}_{i+1} + c^{''}_{i}  &\text{, $0 \ \leq i \ \leq n-1$}       \\
        b^{+}_{i+1} =  a^+_i \overline{a^-_i} \ = \ c^{'}_{i+1}   &\text{, $0 \ \leq i \ \leq n-2$}      \\
    \end{cases}
\end{equation}
${\bf 2^{\grave{e}me} Etape:}$  \\
On a les conditions aux bords suivantes: $d^{'}_{0} \ = \ 1$ ,  $d^{''}_{n} \ = \ 0$. \\
Des \'equations (\ref{L:num16}) et (\ref{L:num20}), on d\'eduit:
\begin{equation}  \label{L:num22}
  d^{'}_{i+1} \ = \ b^+_i \overline{b^-_i} \ = \  c^{'}_{i} \overline{(c^{'}_{i+1} + c^{''}_{i})} , \ \ \ 0 \ \leq \ i \ \leq \ n-1  
\end{equation}
De m\^eme, des \'equations (\ref{L:num16}) et (\ref{L:num20}), on d\'eduit:
\begin{equation}  \label{L:num24}
 d^{''}_{i} \ = \ \overline{b^+_i} b^-_i \ = \  \overline{c^{'}_{i}} (c^{'}_{i+1} + c^{''}_{i}) , \ \ \ 0 \ \leq \ i \ \leq \ n-1  
\end{equation}
Ceci nous permet d\'enoncer la proposition fondamentale suivante:
\begin{proposition}   \label{LL:propos1}
$\forall \ n, i \ \in \ \mathbb{N}$, $n \geq 3$  \\
{\rm Si:}  
\begin{center}
$V_n(i+1)$ {\rm est \'egal \`a} $(c^{'}_{n}, c^{''}_{n}, \dots , c^{'}_{0}, c^{''}_{0} )$ 
\end{center}
{\rm Alors,}  
\begin{center}
$V_n(i+2)$ {\rm est \'egal \`a} $(d^{'}_{n}, d^{''}_{n}, \dots , d^{'}_{0}, d^{''}_{0} )$,  
\end{center}
\[
 {\rm o\grave{u} : \ } 
   \begin{cases}
 d^{'}_{0} =  1  \ \ , \ \ d^{''}_{n} =  0        &\text{}   \notag    \\
 d^{'}_{i+1} \ = \  c^{'}_{i} \overline{(c^{'}_{i+1} + c^{''}_{i})} \ , \ \ \ &\text{$0 \ \leq \ i \ \leq \ n-1$}   \notag  \\
 d^{''}_{i} \ = \ \overline{c^{'}_{i}} (c^{'}_{i+1} + c^{''}_{i}) \ , \ \ \ &\text{$0 \ \leq \ i \ \leq \ n-1$}     \notag     
    \end{cases}
\]
\end{proposition}
{\bf D\'emonstration:}  \\
Le r\'esultat d\'ecoule des \'equations (\ref{L:num22}) et (\ref{L:num24}). \\
$\blacksquare$   \\
Le lemme suivant donne la relation liant $V_n(i+1)$ \`a $U_n(i+1)$.  
\begin{lemma}      \label{LL:propos2}
$\forall \ n, i \ \in \ \mathbb{N}$, $n \geq 3$  \\
{\rm Si:}  \\
\begin{center}
$V_n(i+1)$ {\rm est \'egal \`a} $(c^{'}_{n}, c^{''}_{n}, \dots , c^{'}_{0}, c^{''}_{0} )$  
\end{center}
{\rm Alors,}  \\
\begin{center}
$U_n(i+1)$ {\rm est identique \`a} $((b^{+}_{n-1}, b^{-}_{n-1}) (b^{+}_{n-2}, b^{-}_{n-2}) \dots (b^{+}_{0}, b^{-}_{0}))$,
\end{center}
\[
   {\rm o\grave{u}: }  
   \begin{cases}
 b^{+}_{0} =  1        \notag   &\text{}   \\
 b^{-}_{i} \ = \  c^{'}_{i+1} + c^{''}_{i} \ , \ \ &\text{$0 \ \leq \ i \ \leq \ n-1$}   \notag  \\
 b^{+}_{i+1} \ = \ c^{'}_{i+1} \ , \ \ &\text{$0 \ \leq \ i \ \leq \ n-2$}     \notag     
    \end{cases}
\]
\end{lemma} 
{\bf D\'emonstration:}  \\
Le r\'esultat d\'ecoule de l'\'equation (\ref{L:num20}). \\
$\blacksquare$   \\

\begin{figure}[h]
\begin{center}
\scalebox{.75}{\epsfig{file=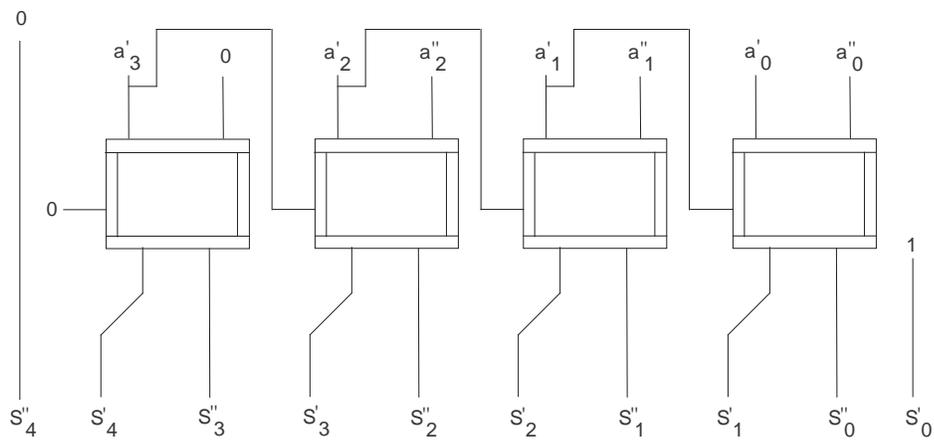}}
\end{center}
\caption{\label{dessin5}{\rm R\'eseau de passage entre deux termes cons\'ecutifs de $V_n$}}
\end{figure}

Le r\'eseau de la figure \ref{dessin5} impl\'emente le passage de $V_n(i+1)$ \`a $V_n(i+2)$.
\newpage

La cellule \'el\'ementaire fonctionne comme indiqu\'e dans la figure \ref{dessin6}:  

\newpage

\begin{figure}[h]
\begin{center}
\scalebox{.75}{\epsfig{file=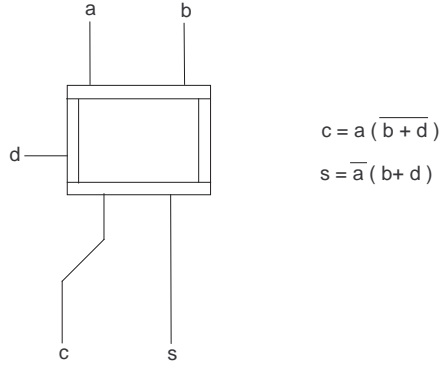}}
\end{center}
\caption{\label{dessin6}{\rm Cellule \'el\'ementaire}}
\end{figure}

\begin{lemma}  \label{LL:lem5}
$\forall \ i, n \ \in \ \mathbb{N}$, $n \geq 3$  \\
{\rm Si:} \\
\begin{center}
$V_n(i+1)$ {\rm est \'egal \`a} $(e^{'}_{n}, e^{''}_{n}, \dots , e^{'}_{0}, e^{''}_{0})$  
\end{center}
{\rm Alors,} \\
\[
\forall \ j \ \in \ \mathbb{N}, \ 1 \ \leq j \ \leq n \ \ \ (e^{'}_{j}, e^{''}_{j-1}) \ne (1, 1) 
\]
\end{lemma} 
{\bf D\'emonstration:}  \\
Supposons que $U_n(i)$ est identique \`a $((a^{+}_{n-1}, a^{-}_{n-1}) (a^{+}_{n-2}, a^{-}_{n-2}), \dots, (a^{+}_{0}, a^{-}_{0}))$, de la
d\'efinition \ref{L:def3}, $V_n(i+1)$ est \'egal \`a: 
\[
 (d^{'}_{n-1}, 0, d^{'}_{n-2}, d^{''}_{n-1}, \dots, d^{'}_{i}, d^{''}_{i+1}, \dots , d^{'}_{0}, d^{''}_{1}, 1, d^{''}_{0})
\]
o\`u
\[
 d^{'}_{j} \ = \ a^{+}_{j} \overline{a^{-}_{j}} \ \ , \ \ d^{''}_{j} \ = \ \overline{a^{+}_{j}} a^{-}_{j}
\]
Par identification, on a:
\[
   \begin{cases}
 e^{'}_{0} =  1 \ , \ \ e^{''}_{n} =  0        \notag   \ \  &\text{}   \\
 e^{'}_{j} \ = \  d^{'}_{j-1} \ = \ a^{+}_{j-1} \overline{a^{-}_{j-1}}  \ \ &\text{, \ $1 \ \leq \ j \ \leq \ n$}    \notag  \\
 e^{''}_{j} \ = \ d^{''}_{j} \ = \ \overline{a^{+}_{j}} a^{-}_{j}  \ \ &\text{, \ $0 \ \leq \ j \ \leq \ n-1$}     \notag     
    \end{cases}
\]
Il s'ensuit: 
\[
   \begin{cases}
e^{'}_{j} \ = \ a^{+}_{j-1} \overline{a^{-}_{j-1}} \ \ &\text{, \ $1 \ \leq \ j \ \leq \ n$}    \notag  \\
e^{''}_{j-1} \ = \ \overline{a^{+}_{j-1}} a^{-}_{j-1} \ \ &\text{, \ $1 \ \leq \ j \ \leq \ n$}     \notag     
    \end{cases}
\]
Il est clair que $(a^{+}_{j-1} \overline{a^{-}_{j-1}} \ \ , \ \ \overline{a^{+}_{j-1}} a^{-}_{j-1}) \ \ne (1, 1)$ pour tout
 $a^{+}_{j-1}, \ a^{-}_{j-1} \ \in \ \{ 0, 1 \}$, ceci implique que $(e^{'}_{j} \ , \ e^{''}_{j-1}) \ \ne \ (1, 1)$. \\
$\blacksquare$  

\section{Simulation de la suite $V_n$ par un r\'eseau neuronal de McCulloch et Pitts}

L'id\'ee de base est de construire un r\'eseau neuronal de McCulloch et Pitts tel qu'au temps $t$, l'\'etat du r\'eseau
 $(x_{2n+2}(t), x_{2n+1}(t), \dots , x_2(t), x_1(t))$ est \'egal \`a $V_n(t+1)$, o\`u $x_i(t)$ est l'\'etat du neurone $i$ au temps $t$. \\

Du point de vue de la suite $V_n$, entre deux tops cons\'ecutifs, les \'etats du r\'eseau repr\'esentent deux termes cons\'ecutifs de la suite.
Tout se passera comme si on reboucle les sorties du r\'eseau de la figure \ref{dessin5} sur ses entr\'ees.

\subsection{Configuration initiale}     

Soit $n$ un entier sup\'erieur \`a $3$, on note $I(n)$ l'ensemble suivant: 
 $I(n) \ = \ \{ j : j \ \in \ \mathbb{N}, 1 \ \leq \ j \ \leq \ 2n+2 \}$, nous consid\'erons un ensemble de $2n+2$ neurones dont l'\'etat
 au temps $0$ est donn\'e par:
\begin{equation}  \label{W:Z0}
   \begin{cases}
x_{2}(0) \ = \ 1  \ \ &\text{ }      \\
x_{i}(0) \ = \ 0  \ \ &\text{si $i \ \in \ I(n) \setminus { 2 }$}         
    \end{cases}
\end{equation}
Du point de vue de la suite $V_n$, l'\'etat du r\'eseau est tel que:
\begin{equation}   \label{L:num26}
(x_{2n+2}(0), x_{2n+1}(0), \dots , x_2(0), x_1(0)) \ {\rm \ est \ \acute{e}gal \ \grave{a} } \ V_n(1).    
\end{equation}

\subsection{Fonction de transition} 

Sous l'hypoth\`ese qu'en entr\'ee du r\'eseau de la figure \ref{dessin5}, on n'ait que les termes de la suite $V_n$. Notre objectif est de 
construire un r\'eseau neuronal qui entre deux tops cons\'ecutifs simule le r\'eseau de la figure \ref{dessin5}. \\

Nous d\'emontrons dans le sous-paragraphe suivant que le r\'eseau neuronal de McCulloch et Pitts dont les fonctions de transition 
locales sont:

\begin{equation}   \label{L:num28}
    \begin{cases}
x_2(t+1)  \ = \ {\bf 1} ( x_2(t) )      &\text{ }      \\
x_{2i+2}(t+1) \ = \ {\bf 1} ( - x_{2i-1}(t) + x_{2i}(t) - x_{2i+2}(t) - 1 ) \ \ &\text{, \ $1 \ \leq \ i \ \leq n$}  \\
x_{2n+1}(t+1)  \ = \ {\bf 1} ( x_{2n+1}(t) - 1 )   &\text{ }   \\
x_{2i-1}(t+1)  \ = \ {\bf 1} ( x_{2i-1}(t) - x_{2i}(t) + x_{2i+2}(t) - 1 ) \ \ &\text{, \ $1 \ \leq \ i \ \leq n$} 
    \end{cases}  
\end{equation}

simule le r\'eseau de la figure \ref{dessin5}. 

\begin{remark}   
{\rm Les neurones $2i+1$, $2i+2$ ($0 \ \leq \ i \ \leq n$) calculent respectivement les termes $s^{''}_{i}$, $s^{'}_{i}$ en sortie du r\'eseau de
la figure \ref{dessin5}. Au cours de l'\'evolution, l'\'etat du neurone $2$ reste \`a $1$ et l'\'etat du neurone $2n+1$ reste \`a $0$.}  
\end{remark}

\subsection{Dynamique du r\'eseau de McCulloch et Pitts}     

Nous \'etudions la dynamique du r\'eseau neuronal de McCulloch et Pitts dont la configuration initiale est 
donn\'ee par l'\'equation (\ref{L:num26}) et les fonctions de transition locales sont d\'efinies par l'\'equation (\ref{L:num28}).

\begin{lemma}   \label{LL:lem6}
{\rm Dans l'\'evolution du r\'eseau neuronal de McCulloch et Pitts:}  \\
{\rm Si:}  
\[
(x_{2n+2}(t), x_{2n+1}(t), \dots , x_2(t), x_1(t)) \ {\rm \ est \ \acute{e}gal \ \grave{a} } \ V_n(p).   
\]
{\rm Alors,}
\[
(x_{2n+2}(t+1), x_{2n+1}(t+1), \dots , x_2(t+1), x_1(t+1)) \ {\rm \ est \ \acute{e}gal \ \grave{a}} \ V_n(p+1).   
\]
\end{lemma}
{\bf D\'emonstration:}  \\
Le Lemme \ref{LL:lem5} nous assure que:
\begin{equation}   \label{L:num30}
 (x_{2i+2}(t), x_{2i-1}(t)) \ \ne \ (1, 1),  \ \ \ 1 \ \leq \ i \ \leq \ n
\end{equation}
Consid\'erons les termes $a$, $b$ et $d$ de la cellule \'el\'ementaire de la figure \ref{dessin6}. Dans le tableau \ref{tabl1}, certaines
entr\'ees ne sont pas prises en compte pour des raisons qui seront donn\'ees par la suite.
\newpage
\begin{table}
\begin{tabular}{|c|c|c|c|c|c|c|}    \hline
 a & b & d & $\overline{a}(b+d)$ & $a \overline{(b+d)}$ & ${\bf 1}(-a+b+d-1)$   &  ${\bf 1}(a-b-d-1)$      \\  \hline
 0 & 0 & 0 &         0           &             0        &          0            &       0                   \\  \hline
 0 & 0 & 1 &         1           &             0        &          1            &       0                   \\  \hline
 0 & 1 & 0 &         1           &             0        &          1            &       0                   \\  \hline
 0 & 1 & 1 &                     &                      &                       &                           \\  \hline
 1 & 0 & 0 &         0           &             1        &          0            &       1                   \\  \hline
 1 & 0 & 1 &         0           &             0        &          0            &       0                   \\  \hline
 1 & 1 & 0 &         0           &             0        &          0            &       0                   \\  \hline
 1 & 1 & 1 &                     &                      &                       &                           \\  \hline
  \end{tabular}
       \caption{Entr\'ees et sorties de la cellule \'el\'ementaire du r\'eseau de la figure  \ref{dessin6} }
	\label{tabl1} 
\end{table}

Identifions $a$, $b$ et $d$ respectivement avec $x_{2i}(t)$, $x_{2i-1}(t)$ et $x_{2i+2}(t)$. En se basant sur 
l'\'equation (\ref{L:num30}), on ne prend pas en compte les entr\'ees $4$ et $8$ de la table \ref{tabl1}.  \\

On v\'erifie ais\'ement gr\^ace \`a la table \ref{tabl1} que: 

\begin{equation}   \label{L:num32}
    \begin{cases}
x_2(t+1)  \ = \  1     &\text{ }      \\
x_{2i+2}(t+1)  \ = \ x_{2i}(t) \overline{ (x_{2i-1}(t) + x_{2i+2}(t) ) } \ \ &\text{, $1 \ \leq \ i \ \leq n$}   \\
x_{2n+1}(t+1)  \ = \ 0  &\text{ }   \\
x_{2i-1}(t+1)  \ = \ \overline{ x_{2i}(t) } (x_{2i-1}(t) + x_{2i+2}(t) ) \ \ &\text{, $1 \ \leq \ i \ \leq n$}   
    \end{cases}
\end{equation}

Identifions $c^{'}_{i}$, $c^{''}_{i}$ ($0 \ \leq \ i \ \leq \ n$) de la Proposition \ref{LL:propos1} respectivement avec
 $x_{2i+2}(t)$ et $x_{2i+1}(t)$. La Proposition \ref{LL:propos1} nous permet de conclure que:
\[
(x_{2n+2}(t+1), x_{2n+1}(t+1), \dots , x_2(t+1), x_1(t+1)) \ {\rm \ est \ \acute{e}gal \ \grave{a} } \ V_n(p+1).   
\]
$\blacksquare$. 
\newpage
\begin{theorem}  \label{LL:lem8}
$\forall \ n \ \in \ \mathbb{N}$, $n \geq 3$, {\rm le r\'eseau neuronal de McCulloch et Pitts:}   \\
\[
x(t+1) \ = \ {\bf 1} ( A x(t) - b )
\]
o\`u $A$ n'est pas sym\'etrique admet $2n+2$ neurones qui d\'ecrivent une \'evolution cyclique de longueur $2^n$.  
\end{theorem}
{\bf D\'emonstration:}  \\
Consid\'erons le r\'eseau neuronal de McCulloch et Pitts dont la configuration initiale est donn\'ee par l'\'equation
 (\ref{L:num26}) et les fonctions de transition locales sont donn\'ees par l'\'equation (\ref{L:num28}). \\
Au temps $t=0$, on a: $(x_{2n+2}(0), x_{2n+1}(0), \dots , x_2(0), x_1(0))$ \ \ est \ \'egal \ \`a \ $V_n(1)$.  \\
Le Lemme \ref{LL:lem6} nous permet de d\'eduire que:
\begin{equation}   \label{L:num34}
\forall \ t \ \in \ \mathbb{N}, \ (x_{2n+2}(t), x_{2n+1}(t), \dots , x_2(t), x_1(t)) \ {\rm \ est \ \acute{e}gal \ \grave{a} } \ V_n(t+1) 
\end{equation}
La d\'efinition \ref{L:def3} et le Lemme \ref{LL:lem4} nous permettent de d\'eduire que $V_n(i+1)$ n'est pas \'egal \`a $V_n(j+1)$, pour
 $0 \ \leq \ i \ < \ j \ \leq 2^n-1$.  \\ 
De l'\'equation (\ref{L:num34}) et de ce qui pr\'ec\'ede, on d\'eduit:
\[
x(t) \ \ne \ x(t') \ , \ \ \ \ \ 0 \ \leq t \ < \ t' \ \leq -1+2^n
\]
De la d\'efinition \ref{L:def3} et du Lemme \ref{LL:lem3}, on d\'eduit que $V_n(1)$ est identique \`a $V_n(1+2^n)$. Ce qui implique que:
 $x(0) \ = \ x(2^n)$. \\
$\blacksquare$  

\newpage

{\bf Exemple 1:} \\

Nous exhibons une \'evolution cyclique de longueur $8$ sur 8 neurones pour illustrer le Th\'eor\`eme \ref{LL:lem8} dans le cas o\`u $n$ est \'egal 
 \`a $3$. 
\begin{table}[h]
  \begin{center}
     \begin{tabular}{|r|c|c|c|c|c|c|c|c|}   \hline
Neurones           &     &     &   &    &   &   &   &         \\
                   &  8  &  7  & 6 &  5 & 4 & 3 & 2 & 1       \\  
 Temps             &     &     &   &    &   &   &   &        \\  \hline
 0                 &  0  &  0  & 0 &  0 & 0 & 0 & 1 & 0       \\  \hline
 1                 &  0  &  0  & 0 &  0 & 1 & 0 & 1 & 0       \\  \hline
 2                 &  0  &  0  & 1 &  0 & 0 & 0 & 1 & 0       \\  \hline
 3                 &  1  &  0  & 0 &  0 & 1 & 1 & 1 & 0       \\  \hline
 4                 &  0  &  0  & 0 &  1 & 0 & 0 & 1 & 0       \\  \hline
 5                 &  0  &  0  & 0 &  1 & 1 & 0 & 1 & 0       \\  \hline
 6                 &  0  &  0  & 1 &  1 & 0 & 0 & 1 & 0       \\  \hline
 7                 &  0  &  0  & 0 &  0 & 1 & 1 & 1 & 0       \\  \hline
 8                 &  0  &  0  & 0 &  0 & 0 & 0 & 1 & 0       \\  \hline 
    \end{tabular}
   \end{center}
	\caption{Evolution Cyclique de longueur 8 sur 8 neurones}
	\label{tabl2} 
\end{table}

{\bf Commentaire 2:} on remarque que les fonctions de transition locales du r\'eseau de neurones de McCulloch et Pitts d\'efinies les
 \'equations (\ref{L:num28}) sont r\'eguli\`eres. 

\section{Evolution pour les r\'eseaux g\'en\'eraux de Caianiello}

L'id\'ee de base est d'exploiter la r\'egularit\'e des fonctions de transition locales
d'une famille d'\'equations neuronales de McCulloch et Pitts de taille $2nk+2$ pour le
simuler \`a l'aide d'un r\'eseau neuronal de Caianiello de taille $2n+2$ et de taille
m\'emoire $k$.  \\
Soit un r\'eseau neuronal de McCulloch et Pitts de taille $2n+2$ dont les fonctions de transition locales
sont:
\begin{equation}   \label{E:pgpp2}
     \begin{cases}
            x_2(t+1) \ = \ {\bf 1} ( x_2(t)   )     &\text{}  \\
            x_{2i+2}(t+1) \ = \  {\bf 1} ( a_1 x_{2i-1}(t) + a_2 x_{2i}(t) + a_3  x_{2i+2}(t) - \theta_1 )  &\text{$1 \ \leq \ i \ \leq n$} 
     \end{cases}
\end{equation}

\begin{equation}   \label{E:pgpp3}
     \begin{cases}
          x_{2n+1}(t+1) \ = \ {\bf 1} ( x_{2n+1}(t) - 1  )     &\text{}  \\
          x_{2i-1}(t+1) \ = \  {\bf 1} ( b_1 x_{2i-1}(t) + b_2 x_{2i}(t) + b_3 x_{2i+2}(t) - \theta_2 )  &\text{$1 \ \leq \ i \ \leq n$} 
     \end{cases}
\end{equation}
et dont l'\'etat initial $(x_{2n+2}(0),  x_{2n+1}(0), \dots , x_2(0) , x_1(0) )$ v\'erifie la contrainte 
\begin{equation}   \label{E:pgpp4}
     \begin{cases}
          x_{2}(0) \ =  \ 1          \\
          x_{2n+1}(0) \ = \ 0    
     \end{cases}
\end{equation}
qui d\'ecrit un transitoire de longueur $T(2n+2)$ et un cycle de longueur $L(2n+2)$. Dans son \'evolution, le r\'eseau neuronal
de McCulloch et Pitts de taille $2n+2$ d\'efini par les \'equations neuronales (\ref{E:pgpp2}) et (\ref{E:pgpp3}) d\'ecrit une suite:
\[
W_{n}(0), W_{n}(1),  W_{n}(2) , \cdots ,  W_{n}(-1+T(2n+2)+L(2n+2)) , \cdots 
\]
o\`u:
\[
W_{n}(i) \ = \ (x_{2n+2}(i), x_{2n+1}(i), \dots , x_2(i), x_1(i)).
\]

Le graphe de d\'ependance des fonctions de transition locales des neurones $2$, $2i+2(1 \ \leq \ i \ \leq \ n)$,
 $2i-1(1 \ \leq \ i \ \leq \ n)$ et $2n+1$ d\'efinies d'une part par les \'equations (\ref{L:num28}) et d'autre part
par les \'equations (\ref{E:pgpp2}), (\ref{E:pgpp3}) est:
\newpage
\begin{figure}[h]
\begin{center}
\scalebox{.75}{\epsfig {file = 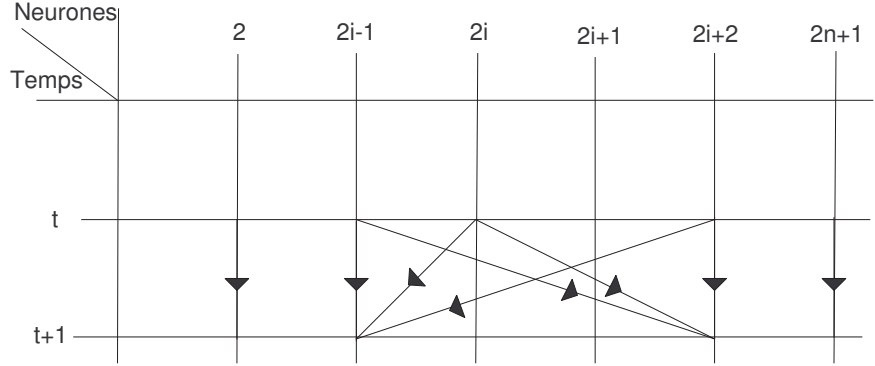  }}
\end{center}
\caption{graphe de d\'ependance des fonctions de transition locales}
\end{figure}

L'id\'ee est d'exploiter la r\'egularit\'e des fonctions de transition locales du r\'eseau neuronal de McCulloch et Pitts d\'efini en
 (\ref{E:pgpp2}) et en (\ref{E:pgpp3})
pour construire un r\'eseau de Caianiello de taille $2n+2$ et de m\'emoire $k$ qui simule en $k$ it\'erations une it\'eration de
l'\'equation neuronale de McCulloch et Pitts qui impl\'emente le passage entre deux termes cons\'ecutifs de la suite $W_{nk}$. \\
Le r\'eseau neuronal de McCulloch et Pitts qui impl\'emente la suite $W_{nk}$ a pour \'etat initial 
$(x_{2nk+2}(0), x_{2nk+1}(0), \dots , x_2(0), x_1(0))$  avec la contrainte: 
\begin{equation}   \label{E:pgpp5}
     \begin{cases}
          x_{2}(0) \ = \ 1    &\text{}  \\
          x_{2nk+1}(0) \ = \  0  &\text{ } 
     \end{cases}
\end{equation} 
et dont les fonctions de transition locales sont:
\begin{equation}   \label{E:pgpps}
  \begin{cases}   
x_2(t+1) \ = \ {\bf 1} ( x_2(t) )      &\text{}  \\
x_{2i+2}(t+1) \ = \ {\bf 1} ( a_1 x_{2i-1}(t) + a_2 x_{2i}(t) + a_3  x_{2i+2}(t) - \theta_1 ) \ &\text{$1 \ \leq \ i \ \leq nk$} \\
x_{2nk+1}(t+1) \ = \ {\bf 1} ( x_{2nk+1}(t) - 1 )        &\text{}   \\
x_{2i-1}(t+1) \ = \  {\bf 1} ( b_1 x_{2i-1}(t) + b_2 x_{2i}(t) + b_3 x_{2i+2}(t) - \theta_2 ) \ &\text{$1 \ \leq \ i \ \leq nk$}
  \end{cases} 
\end{equation}

\begin{remark}   \label{lr:r3}
{\rm Dans le r\'eseau neuronal de Caianiello d\'efini par l'\'equation (\ref{E:pgpps}), en affectant \`a:}
\begin{align}  
&a_1 \ = \ -1 ; \ a_2 \ = \ 1  ; \ a_3 \ = \ -1 ; \ \theta_1 \ = \ 1      \label{ry:yr2}  \\ 
&b_1 \ = \  1 ; \ b_2 \ = \ -1 ; \ b_3 \ = \  1 ; \ \theta_2 \ = \ 1 \    \label{ry:yr3}  \\
&x_1(0) \ = \ 0 ; \ x_i(0) \ = \ 0 \ , \ 3 \ \leq \ i \ \leq 2nk \ \ ; \ \ x_{2nk+2}(0) \ = \ 0  \label{ry:yr4}
\end{align}
{\rm on retrouve le r\'eseau neuronal de McCulloch et Pitts de taille $2nk+2$ qui simule la suite $V_{nk}$. 
De plus, en affectant aux $a_i$, $b_i$ et $\theta_i$ les valeurs donn\'ees par les \'equations (\ref{ry:yr2}), (\ref{ry:yr3}), et (\ref{ry:yr4})
 et en imposant $k=1$, on retrouve le r\'eseau neuronal de McCulloch et Pitts d\'efini par l'\'equation (\ref{L:num28}).}
\end{remark}

La simulation en $k$ it\'erations par un r\'eseau neuronal de Caianiello constitu\'e de $2n+2$ neurones de taille m\'emoire $k$ d'une
it\'eration de l'\'equation (\ref{E:pgpps}) se d\'eroule suivant les r\`egles ci-dessous: \\
{\bf R\`egle 1:} \\
L'\'etat du neurone $j$ ($1 \ \leq j \ \leq \ 2n$) du r\'eseau neuronal de Caianiello au temps $tk+i$ ($0 \leq i \leq k-1$) est \'egal 
 \`a l'\'etat du neurone $2ni+j$ de l'\'equation (\ref{E:pgpps}) au temps $t$, en d'autre termes:
\[
y_{j}(tk+i) \ = \ x_{2ni+j}(t)
\]
{\bf R\`egle 2:} \ \ \ au top (t+1)k-1   \\
L'\'etat du neurone $2n+1$ du r\'eseau neuronal de Caianiello au temps $(t+1)k-1$ est \'egal \`a l'\'etat du neurone $2nk+1$ de 
 l'\'equation (\ref{E:pgpps}) au temps $t$, c'est-\`a-dire:
\[
y_{2n+1}((t+1)k-1) \ = \ x_{2nk+1}(t)
\]
L'\'etat du neurone $2n+2$ du r\'eseau neuronal de Caianiello au temps $(t+1)k-1$ est \'egal \`a l'\'etat du neurone $2nk+2$ de 
 l'\'equation (\ref{E:pgpps}) au temps $t$, ce qui entra\^{i}ne:
\[
y_{2n+2}((t+1)k-1) \ = \ x_{2nk+2}(t)
\]
Etant donn\'e que l'\'etat du neurone $2n+2$ du r\'eseau neuronal de Caianiello au temps $(t+1)k-1$ est $x_{2nk+2}(t)$ et que par souci
de r\'egularit\'e, on aimerait que l'\'etat de tout neurone $i$, $i \ \in \ \{1, 2, \dots, 2n, 2n+2 \}$ \`a deux tops 
cons\'ecutifs de l'intervalle $[ tk \ , \ (t+1)k-1 ]$ soit respectivement $x_{s}(t)$ et  $x_{s'}(t)$, $s$ et $s'$ v\'erifiant la
relation suivante $s \ - \ s' \ = \ 2n$. On introduit la r\`egle suivante: 
\newpage
{\bf R\`egle 3:}   \\
L'\'etat du neurone $2n+2$ du r\'eseau neuronal de Caianiello au temps $tk+i$ est \'egal \`a l'\'etat du neurone $2n(i+1)+2$ de l'\'equation 
 (\ref{E:pgpps}) au temps $t$, en d'autres termes:
\[
y_{2n+2}(tk+i) \ = \ x_{2n(i+1)+2}(t)
\]
On remarque \'egalement que l'\'etat du neurone d'indice $2$ de l'\'equation (\ref{E:pgpps}) est constant au cours de l'\'evolution
 alors que l'\'etat du neurone d'indice $2ni+2 \ (0 \ < \ i \ \leq k)$ du r\'eseau neuronal (\ref{E:pgpps}) n'est pas constant    
au cours de l'\'evolution. Etant donn\'e que c'est le neurone d'indice $2$ du r\'eseau de Caianiello qui simule au top $tk+i \
(0 \ \leq \ i \ \leq k-1)$ le neurone d'indice $2ni+2$ du r\'eseau neuronal (\ref{E:pgpps}), on introduit la r\`egle 4 suivante: \\
{\bf R\`egle 4:}  \\
Le neurone $2n+1$ du r\'eseau neuronal de Caianiello d\'elivre $1$ au top $tk$ et $0$ au top $tk+i$ ($1 \ \leq \ i \ \leq k-1$). \\
{\bf Commentaire 3:}  \\
L'\'etat du neurone $2n+1$ du r\'eseau neuronal de Caianiello au top $t$ permet au neurone $2$ du r\'eseau neuronal de Caianiello de savoir
s'il simule au top $t+1$ le neurone $2$ du r\'eseau neuronal de McCulloch et Pitts ou non. \\

Sch\'ematiquement, l'\'etat du r\'eseau neuronal de Caianiello en fonction de l'\'etat du r\'eseau neuronal (\ref{E:pgpps}) de McCulloch 
et Pitts aux temps $ak, \ ak+1, \ ak+2, \dots , (a+1)k-1$ est illustr\'e dans le tableau \ref{tabl3}:
\newpage
\begin{table}[h]
	\begin{center}
		\begin{tabular}{|c|c|c|c|c|c|c|}  \hline
		 Temps  & & & & & & \\
		   & $ak$ & $a k+1$ & $\cdots$ & $a k+i$ & $\cdots$ & $ \left(a+1\right) k-1$ \\
		  Neurones & & & & & & \\ \hline
		 $y_1$ & $x_1\left(a\right)$ & $x_{2n+1}\left(a\right)$ & & $x_{2ni+1}\left(a\right)$ & & $x_{2n\left(k-1\right)+1}\left(a\right)$ \\ \hline
		 $\vdots$ & $\vdots$ & $\vdots$ & & & & \\ \hline
		 $y_j$ & $x_j\left(a\right)$ & $x_{2n+j}\left(a\right)$ & & $x_{2ni+j}\left(a\right)$ & & $x_{2n\left(k-1\right)+j}\left(a\right)$ \\ \hline
		 $\vdots$  & $\vdots$ & $\vdots$ & & & & \\ \hline
		 $y_{2n}$ & $x_{2n}\left(a\right)$ & $x_{4n}\left(a\right)$ & & $x_{2n\left(i+1\right)}\left(a\right)$ & & $x_{2nk}\left(a\right)$ \\ \hline
		 $y_{2n+1}$ & 1 & 0 & & 0 & & 0 \\ \hline
		 $y_{2n+2}$ & $x_{2n+2}\left(a\right)$ & $x_{4n+2}\left(a\right)$ & & $x_{2n\left(i+1\right)+2}\left(a\right)$ & & $x_{2nk+2}\left(a\right)$ \\ \hline

		\end{tabular}
	\end{center}
\caption{Simulation}
    	\label{tabl3} 
\end{table}

S\'emantiquement $(y_{2n+2}((a+1)k-1),\dots, y_{1}((a+1)k-1), \dots , y_{2n}(ak+i), \dots, y_{1}(ak+i), \dots, y_{2n}(ak), \dots, y_{1}(ak))$
repr\'esente $W_{nk}(a)$. \\
Notre but est de d\'emontrer que le r\'eseau neuronal de Caianiello dont l'\'etat initial est:
\begin{equation} \label{E:pg0g}
 \begin{cases}
     y_{j}(i) \ = \ x_{2ni+j}(0)  \ &\text{$1 \ \leq \ i \ \leq 2n$ \ et \ $0 \ \leq i \ \leq k-1$}  \\
     y_{2n+1}(0) \ = \ 1    \ \ \ \ &\text{}    \\
     y_{2n+1}(i) \ = \ 0          \ &\text{$1 \ \leq i \ \leq k-1$} \\
     y_{2n+2}(i) \ = \ x_{2n(i+1)+2}(0)  \ &\text{$0 \ \leq i \ \leq k-1$}
 \end{cases} 
\end{equation}
et les fonctions de transition locales sont:
\begin{equation}   \label{E:pgg}
  \begin{cases}   
y_2(t+1) \ = \ {\bf 1} (y_{2n+1}(t+1-k) + y_{2n+2}(t) - 1)     &\text{ }  \\
y_{2i+2}(t+1) \ = \  {\bf 1} ( a_1 y_{2i-1}(t+1-k) + a_2 y_{2i}(t+1-k) + a_3 y_{2i+2}(t+1-k) - \theta_1 ) 
                                 \  &\text{$1 \ \leq \ i \ \leq n$} \\
y_{2n+1}(t+1) \ = \ {\bf 1} ( y_{2n+1}(t+1-k) - 1 )      &\text{ }   \\
y_{2i-1}(t+1) \ = \ {\bf 1} ( b_1 y_{2i-1}(t+1-k) + b_2 y_{2i}(t+1-k) + b_3 y_{2i+2}(t+1-k) - \theta_2 ) 
                       \ &\text{$1 \ \leq \ i \ \leq n$}
  \end{cases} 
\end{equation}
simule le r\'eseau neuronal (\ref{E:pgpps}) de McCulloch et Pitts. La proposition ci-dessous \'etablit le lien entre les it\'erations
des r\'eseaux (\ref{E:pgpps}) et (\ref{E:pgg}). 
\begin{proposition}  \label{ch4p:pp00}    
{\rm Dans l'\'evolution du r\'eseau neuronal (\ref{E:pgg}) de Caianiello, soit} $e \ \in \ \mathbb{N}$: \\
{\rm Si:} \\
$a)$ \quad $y_{j}(ek+i) \ = \ x_{2ni+j}(e)$, \ $\forall \ i, j \ \in \mathbb{N}$ {\rm \ tel \ que}  $0 \ \leq \ i \ \leq k-1$, \\
\phantom{ salut rene     ndoundam               bonjour                 } $1 \ \leq \ j \ \leq 2n+2$ \ {\rm et} \ $j \ne 2n+1$     \\
$b)$ 
   \begin{align}   \notag
       y_{2n+1}(ek) \ &= \ 1         \notag   \\
       y_{2n+1}(ek+j) \ &= \ 0 \ , \ \ \ 1 \ \leq j \ \leq k-1   \notag   
   \end{align}
{\rm Alors} \\
$c)$ \quad $y_{j}((e+1)k+i) \ = \ x_{2ni+j}(e+1)$ , \ \ $\forall \ i, j \ \in \mathbb{N}$ {\rm \ tel \ que} \quad $0 \ \leq \ i \ \leq k-1$, \\
\phantom{                      salut rene     ndoundam               bonjour                 } $1 \ \leq \ j \ \leq 2n+2$ \ {\rm et}
\ $j \ne 2n+1$  \\
$d)$ 
   \begin{align}   \notag
       y_{2n+1}((e+1)k) \ &= \ 1         \notag   \\
       y_{2n+1}((e+1)k+j) \ &= \ 0 \ \ {\rm pour} \ j \ {\rm tel \ que \ } \ 1 \ \leq j \ \leq k-1   \notag   
   \end{align}
\end{proposition}
{\bf D\'emonstration:}  \\
La d\'emonstration se fera en deux \'etapes, la premi\`ere \'etape et la deuxi\`eme \'etape sont consacr\'ees respectivement aux neurones
d'indices pairs et impairs. \\
{\bf  Premi\`ere \'etape:} les neurones d'indices pairs  \\
$1)$  le neurone d'indice $2j+2$,    $1 \ \leq j \ \leq n$     \\
La d\'emonstration s'appuie sur les hypoth\`eses de la Proposition \ref{ch4p:pp00}. Au top $t = \ (e+1)k+i$, l'\'etat du neurone $2j+2$ est:
\begin{align}
y_{2j+2}((e+1)k+i) \ &= \ {\bf 1} ( a_1 y_{2j-1}(ek+i) + a_2 y_{2j}(ek+i) + a_3 y_{2j+2}(ek+i) - \theta_1  )  \notag    \\
   &= \ {\bf 1} ( a_1 x_{2ni+2j-1}(e) + a_2 x_{2ni+2j}(e) + a_3 x_{2ni+2j+2}(e) - \theta_1 ) \ \ \text{par hypoth\`ese} \notag \\
   &= \ x_{2ni+2j+2}(e+1) \ \ \ \text{par application de l'\'equation (\ref{E:pgpps})}    \notag  
\end{align}
$2)$  le neurone d'indice $2$  \\
Consid\'erons le top $t=(e+1)k+i$, la d\'emonstration se fera de proche en proche sur $i$. \\
{\bf Rang 0}: \ \ au top $t=(e+1)k$   \\   
\begin{align}
y_{2}((e+1)k) \ &= \ {\bf 1} ( y_{2n+1}(ek) +  y_{2n+2}(ek+k-1) - 1  )  &\text{}  \notag    \\
   &= \ {\bf 1} ( 1 + x_{2nk+2}(e) - 1 )  &\text{par hypoth\`ese on a $y_{2n+1}(ek) \ = \ 1$} \notag \\
   &= \ 1    &\text{ }  &\text{ }   &\text{}   \notag  \\
   &= x_{2}(e+1)  &\text{car $x_2(t) \ = \ 1, \ \ \forall \ t \ \in \ \mathbb{N}$}    \notag  
\end{align}
{\bf Hypoth\`ese de r\'ecurrence}: \ on suppose la propri\'et\'e vraie au rang $i-1$  \\
{\bf Rang i}: au top $t \ = \ (a+1)k+i, \ \ \ 1 \ \leq \ i \ \leq \ k-1$   
\begin{align}
y_{2}((e+1)k+i) \ &= \ {\bf 1} ( y_{2n+1}(ek+i) + y_{2n+2}((e+1)k+i-1) - 1 )  \notag    \\
   &= \ {\bf 1} ( y_{2n+2}((e+1)k+i-1)  - 1 ) \ \ \text{par hypoth\`ese $y_{2n+1}(ek+i) \ = \ 0$}   \notag \\
 &= \ {\bf 1} (  x_{2ni+2}(e+1)  - 1 ) \ \ \text{de $1)$, on a $y_{2n+2}((e+1)k+i-1) \ = \ x_{2ni+2}(e+1)$}   \notag \\
   &= \  x_{2ni+2}(e+1)   \notag
\end{align}

{\bf Deuxi\`eme \'etape:} les neurones d'indices impairs \\

$3)$  le neurone d'indice $2j-1, \ \ 1 \ \leq j \ \leq \ n$   \\
La d\'emonstration s'appuie sur les hypoth\`eses de la Proposition \ref{ch4p:pp00}. Au top \\  $t \ = \ (e+1)k+i$, l'\'etat du neurone $2j-1$ est: 
\begin{align}
y_{2j-1}((e+1)k+i) \ &= \ {\bf 1} ( b_1 y_{2j-1}(ek+i) + b_2 y_{2j}(ek+i) + b_3 y_{2j+2}(ek+i)  - \theta_2 )   \notag    \\
 &= \ {\bf 1} ( b_1 x_{2ni+2j-1}(e) + b_2 x_{2ni+2j}(e) + b_3 x_{2ni+2j+2}(e) - \theta_2 ) \ \ \text{par hypoth\`ese} \notag \\
   &= \  x_{2ni+2j-1}(e+1) \ \ \ \text{par application de l'\'equation (\ref{E:pgpps})}   \notag 
\end{align}
$4)$  le neurone d'indice $2n+1$  \\
La d\'emonstration se fera en se basant sur les hypoth\`eses de la proposition. Au top $t \ = \ (a+1)k+i$, l'\'etat du neurone $2n+1$
est:
\begin{align}
y_{2n+1}((e+1)k+i) \ &= \ {\bf 1} ( y_{2n+1}(ek+i) - 1 )   \notag    \\
 &= \  y_{2n+1}(ek+i)   \notag 
\end{align}
En se basant sur les hypoth\`eses de la Proposition \ref{ch4p:pp00} (en particulier sur le $b°)$), on d\'eduit le r\'esultat. \\
$\blacksquare$  
\begin{theorem}  \label{ch4p:pp01} .   \\ 
{\rm Si} \\
 {\it RMC} \ {\rm est un r\'eseau neuronal de McCulloch et Pitts de taille} $2nk+2$ {\rm dont les fonctions de transition locales
 sont donn\'ees par l'\'equation (\ref{E:pgpps}) et l'\'etat initial v\'erifie la contrainte de l'\'equation (\ref{E:pgpp5}) qui d\'ecrit un
 transitoire de longueur} $T(2nk+2)$ {\rm et un cycle de longueur} $L(2nk+2)$  \\
{\rm Alors,}   \\ 
{\rm il existe un r\'eseau neuronal de Caianiello de taille} $2n+2$ {\rm et de m\'emoire}
$k$ {\rm qui simule} {\it RMC} {\rm et qui d\'ecrit un transitoire de longueur} $k \times T(2nk+2)$ {\rm et un cycle de longueur} $k \times L(2nk+2)$.
\end{theorem}
{\bf D\'emonstration} La preuve est une cons\'equence de la Proposition \ref{ch4p:pp00}.  \\
$\blacksquare$ \\
Le Th\'eor\`eme \ref{ch4p:pp01} nous permet de retrouver l'un des r\'esultats de l'article \cite{NT 00c}:
\begin{corollary} \cite{NT 00c} .   \\
{\rm Il existe un r\'eseau neuronal de Caianiello constitu\'e de} $2n+2$ {\rm neurones de taille m\'emoire} $k$ {\rm qui d\'ecrit un cycle
de longueur} $k2^{nk}$
\end{corollary}
{\bf D\'emonstration}  \\
Du Th\'eor\`eme \ref{LL:lem8} et de la Remarque \ref{lr:r3}, on d\'eduit qu'il existe une instance {\it RMC} du r\'eseau neuronal (\ref{E:pgpps})
de McCulloch et Pitts qui simule la suite $V_{nk}$. Par application du Th\'eor\`eme \ref{ch4p:pp01} \`a l'instance {\it RMC} du r\'eseau neuronal
 (\ref{E:pgpps}) de McCulloch et Pitts qui simule la suite $V_{nk}$, on d\'eduit le r\'esultat.
$\blacksquare$  

\section{Conclusion}
L'interdiction de certains codes a \'et\'e utilis\'e pour construire tr\`es simplement, \`a partir de compteurs dans des syst\`emes arithm\'etiques
redondants o\`u l'addition se fait sans propagation de retenue, des r\'eseaux de McCulloch et Pitts qui g\'en\`erent des cycles de longueur
 exponentielle. La contribution est la m\'ethode de simulation d'une famille d'\'equations neuronales de McCulloch et Pitts de taille $2nk+2$ dont
 les fonctions de transition locales sont r\'eguli\`eres par une famille de r\'eseaux neuronaux de Caianiello de taille $2n+2$ et de taille
 m\'emoire $k$. Cette technique vient enrichir la gamme des constructions combinatoires qui ont d\'ej\`a montr\'e leur puissance dans le
 pass\'e \cite{Tch 86, CTT 92, Mat 95, NT 03, NT 04}.


\begin{thebibliography}{2}
\bibitem{CL 66} E. R. Caianiello and A. De Luca,
{"Decision Equation for Binary Systems : Applications to Neuronal
Behavior,"} {\it Kybernetic}, {\bf 3}(1966) 33-40
\bibitem{CMG 88} M. Cosnard, D. Moumida, E. Goles and T.de.St. Pierre,
{"Dynamical Behavior of a Neural Automaton with Memory,"}  {\it
Complex Systems,} {\bf 2}, (1988), 161-176.
\bibitem{CTT 92} M. Cosnard, M. Tchuente, G. Tindo,
{"Sequences generated by neuronal automata with memory,"}  {\it
Complex Systems,} {\bf 6}, (1992), 13-20.
\bibitem{DM 91} J. Duprat, J. M. Muller, {"Ecrire les nombres autrement pour calculer plus vite"},{\it Technique et Science Informatique},
Vol. 10 N°{\bf (3)} pp 211-224.
\bibitem{MP 43} W. S. Mc Culloch and W. Pitts, {"A Logical Calculus of the
Ideas Immanent in Nervous Activity"}, {\it Bulletin of Mathematical
Biophysics,} {\bf 5} (1943) 115-133.
\bibitem{Gol 80} E. Goles, {"Comportement oscillatoire d'une famille d'automates cellulaires non uniformes"}, {\it Doctoral Dissertation,}
University of Grenoble, 1980.
\bibitem{GO 81} E. Goles, J. Olivos, {"Comportement P\'eriodique des Fonctions \`a seuil Binaires et Applications",}
 {\it Disc. App. Math.}, {\bf 3}, 1981, 95-105.  
\bibitem{Gol 82} E. Goles,  {"Fixed Point Behaviour of Threshold Functions on a finite set,"} {\it
SIAM J. on Alg. and Disc Methods,} {\bf 3(4)}(1982) pages 529-531. 
\bibitem{GT 83} E. Goles and M. Tchuente, {"Iterative Behaviour of
Generalized Majority Functions",} {\it Mathematical Social Sciences}, {\bf 4},
1983, 197-204.
\bibitem{Gol 85} E. Goles, {"Dynamical Behavior of Neural Networks",} {\it SIAM J. Disc. Alg. Methods}, {\bf 6},
1985, 749-754.
\bibitem{GM 89} E. Goles and S. Mart\'{\i}nez, {"Exponential Transient Classes of
Symmetric Neural Networks for Synchronous and Sequential
Updating"},{\it Complex Systems}, {\bf 3(1989)} 589-597.
\bibitem{GM 90} E. Goles and S. Mart\'{\i}nez,
{" Neural and Automata Networks, Dynamical Behaviour and Applications"}
(Norwell, MA, Kluwer Academic Publishers, 1990)
\bibitem{GHM 89} A. Guyot, Y. Herreros, J.M.Muller, {"Janus: an on-line Multiplier-Divider for Manipulating Large Numbers"},{\it Proceedings 
IEEE $9^{th}$ Symposium on Computer Arithmetic,}, Santa-Monica, Sept 1989. 
\bibitem{Mou 89} D. Moumida, {"Contribution \`a l'\'etude de la Dynamique d'un Automate \`a m\'emoire"}, {\it Doctoral Dissertation,}
University of Grenoble, 1989.
\bibitem{Mat 95} M. Matamala, {"Recursive Construction of Periodic Steady
State for Neural Networks"}, {\it Theoretical Computer Science,}
{\bf 143(2)} (1995) 251-267.
\bibitem{Nd 95} R. Ndoundam  {"Analyse et Synth\`ese de Certains R\'eseaux d'Automates"}, {\it Th\`ese de Doctorat de 3eme Cycle,},
Universit\'e de Yaound\'e I, 1995. 
\bibitem{NM 00b} R. Ndoundam and M. Matamala, {"No Polynomial Bound for the
Period of Neuronal Automata with Memory With  Inhibitory
Memory"}, {\it Complex Systems}, {\bf 12(2000)}, 391-397.
\bibitem{NT 00c} R. Ndoundam and M. Tchuente, {"Cycles exponentiels des r\'eseaux de Caianiello et compteurs
 en arithm\'etique redondante"}, {\it Technique et Science Informatiques}, Vol {\bf 19}, N° 7/2000 pages 985-1008.
\bibitem{NT 03} R. Ndoundam and M. Tchuente, {"Exponential transient length generated by a neuronal recurrence 
 equation"}, {\it Theoretical Computer Science}, {\bf 306(2003)}, 513-533.
\bibitem{NT 04} R. Ndoundam and M. Tchuente, {"Exponential Period of Neuronal Recurrence Automata
 with Excitatory Memory"}, {\it Complex Systems}, {\bf 15(2004)}, 79-88.
\bibitem{Tch 86} M. Tchuente {"Sequential Simulation of Parallel
Iterations and Applications"},{\it Theoretical Computer Science}, {\bf vol.
48}, 1986, p. 135-144.
\bibitem{TT 93} M. Tchuente and G. Tindo, {"Suites g\'en\'er\'ees par une
\'equation  neuronale \`a m\'emoire"},{\it Comptes Rendus de l'Acad\'emie des
Sciences}, {\bf tome 317, Serie I},  625-630, 1993.
\end{thebibliography}
\end{document}